\documentclass[conference]{IEEEtran}
\usepackage{cite}
\usepackage{amsmath,amssymb,amsfonts}
\usepackage{algorithm}
\usepackage{algpseudocode}
\usepackage{textcomp}
\usepackage{xcolor}

\providecommand{\doi}[1]{doi: {\footnotesize \href{http://dx.doi.org/#1}{\path{#1}}}}

\usepackage[pdftex=true,breaklinks=true,hidelinks=true,colorlinks=true,citecolor=blue]{hyperref}

\usepackage{graphicx}
\graphicspath{{pdf/}{photo/}{figures/}}
\DeclareGraphicsExtensions{.eps,.jpeg,.jpg,.png,.pdf}

\usepackage{subcaption}
\usepackage[]{caption}

\usepackage{fancyhdr}
\usepackage{booktabs}

\def\BibTeX{{\rm B\kern-.05em{\sc i\kern-.025em b}\kern-.08em
    T\kern-.1667em\lower.7ex\hbox{E}\kern-.125emX}}
\begin{document}

\makeatletter
\fancypagestyle{plain}{
    \fancyhf{}
    \fancyhead[C]{\emph{Preprint submitted to 15\textsuperscript{th} Int. Conf. on Information Society and Technology (ICIST), Kopaonik, Serbia, 9-12 March 2025}}
    \fancyfoot[C]{\thepage}
     \renewcommand{\headrulewidth}{0pt}
}
\makeatother

\makeatletter
\fancypagestyle{custom}{
    \fancyhf{}
    \fancyfoot[C]{\thepage}
    \renewcommand{\headrulewidth}{0pt}
}
\makeatother

\title{Named Entity Recognition for Serbian Legal Documents: Design, Methodology and Dataset Development\\
%{\footnotesize \textsuperscript{*}Note: Sub-titles are not captured for https://ieeexplore.ieee.org  and
%should not be used}
\thanks{Identify applicable funding agency here. If none, delete this.}
}

\author{\IEEEauthorblockN{Vladimir Kalušev}
\IEEEauthorblockA{\textit{Visual Computing \& Perception Group} \\
\textit{The Institute for Artificial Intelligence} \\ \textit{Research and Development of Serbia}\\
Novi Sad, Republic of Serbia \\
vladimir.kalusev@ivi.ac.rs}
\and
\IEEEauthorblockN{Branko Brkljač}
\IEEEauthorblockA{\textit{Dept. of Power, Electronic and Telecommunication Engineering} \\
\textit{Faculty of Technical Sciences, University of Novi Sad}\\
Novi Sad, Republic of Serbia \\
brkljacb@uns.ac.rs}
}

\maketitle
\thispagestyle{plain} % Apply the defined page style to the first page

\begin{abstract}
Recent advancements in the field of natural language processing (NLP) and especially large language models (LLMs) and their numerous applications have brought research attention to design of different document processing tools and enhancements in the process of document archiving, search and retrieval. Domain of official, legal documents is especially interesting due to vast amount of data generated on the daily basis, as well as the significant community of interested practitioners (lawyers, law offices, administrative workers, state institutions and citizens). Providing efficient ways for automation of everyday work involving legal documents is therefore expected to have significant impact in different fields. In this work we present one LLM based solution for Named Entity Recognition (NER) in the case of legal documents written in Serbian language. It leverages on the pre-trained bidirectional encoder representations from transformers (BERT), which had been carefully adapted to the specific task of identifying and classifying specific data points from textual content. Besides novel dataset development for Serbian language (involving public court rulings), presented system design and applied methodology, the paper also discusses achieved performance metrics and their implications for objective assessment of the proposed solution. Performed cross-validation tests on the created manually labeled dataset with mean $F_1$ score of 0.96 and additional results on the examples of intentionally modified text inputs confirm applicability of the proposed system design and robustness of the developed NER solution.
\end{abstract}


\section{Introduction}
Named Entity Recogniton (NER) represents identification and classification of named entities in certain text or document, where named entities are typically noun phrases or predefined categories that refer to some specific object, person, places, dates, or other domain specific entities \cite{li2020survey}. These tools are often used for preprocessing or analysis of text documents for the purpose of information extraction, document tagging, retrieval or search. Thus, the possible use cases for NER tools are very diverse and depend on specific text domain and application. In the context of legal documents, possibility to automatically extract structured information about involved parties (places, reference numbers, court names, dates, laws, official gazette, money amounts, etc.) allows precise archiving, design of efficient search engines and question answering (QA) tools. Since such information is usually the most significant in the document, extraction of key entity values also makes the text summarization, document classification and sentiment analysis easier to perform. Some of the examples of contemporary NER tools specifically targeting legal documents are described in \cite{ccetindaug2023named} for Turkish, \cite{darji2023german} for German, and in \cite{chalkidis2020legal, jin2023teamshakespeare, rajamanickam2024improving} for English language, demonstrating significant interest in this application area. Since Serbian language is still considered as low-resource in context of LLM development and different downstream applications. Thus, NER tools for legal documents in Serbian are still rare and uncommon among practitioners.

In order to overcome existing challenges,  and contribute towards democratization of LLM technology and its proliferation among Serbian speaking community, we demonstrate design and development of one specific NLP downstream application that leverages on fine tuning of pre-trained model (PTM). Proposed NER for Serbian legal documents leverages on the BERT type PTM \cite{kenton2019bert}, which was previously developed for Serbian and other south Slavic languages by \cite{ljubevsic2021bertic}. Presented solution and reported results confirm applicability of proposed design for PTM task adaptation in the case of low-resource languages, and especially in the domain of official legal texts. Therefore it is expected that presented methodology will motivate further development of similar language tools for Serbian language. According to the conducted experiments on the annotated dataset consisting of 75 unique appellate court rulings from standard legal practice, the proposed NER model achieves mean recognition accuracy of 0.99 and individual $F_1$ measures in the range between 0.89 and 0.99 over 15 NER categories corresponding to 8 unique named entity (NE) types characteristic for such legal documents. Besides cross-validation, additional tests involving noised textual data confirm robustness of NER model and its applicability for the specific task. The \textbf{NER4Legal\_SRB} model parameters and proposed dataset are freely available in the following repository: {\url{https://huggingface.co/kalusev/NER4Legal_SRB}}.

\newpage
\pagestyle{custom}

The rest of the paper is organized as follows, in Section~\ref{Related work} we provide an overview of NER design principles and contemporary approaches involving PTM and different NLP representation learning techniques. In Section~\ref{Methods and dataset} are described labeling process and created legal texts dataset, as well as NER model architecture and PTM adaptation to downstream NER task by low-resource fine-tuning over developed dataset. Section~\ref{Results and discussion} presents experimental results and their discussion. Finally, Section~\ref{Conclusion} indicates directions for future work.

\section{Related work}
\label{Related work}

There are different design approaches for NER tools, which are trying to capture intricate language context in order to extract and recognize specific entity values. Entity meaning is usually tied to its surrounding context, which makes the rule based \cite{alfred2014malay, chiticariu2013rule}, and dictionary based methods \cite{hanisch2005prominer} less favourable in contrast to machine learning (ML) based approaches \cite{geng2017clinical, lee2020biobert, chalkidis2020legal} or hybrid architectures \cite{petasis2001using, nastou2024improving}.

Traditional NER methods were typically relying on handcrafted features capturing short-distance relations between the words in the sequence, and lacking the ability to consider bidirectional word relationships. As a result, they often fell short when dealing with complex linguistic scenarios where an entity's meaning reflects the surrounding context. Supervised NER solves a multi-class classification problem or a sequence labeling task, where each of the labeled training samples is represented by the corresponding feature vector, and the corresponding ML model is used to recognize new samples from the text. Depending on the classification model, there had been various learning approaches mainly based on sequence modelling capabilities of Hidden Markov Models (HMMs) \cite{bikel1997nymble}, Conditional Random Fields (CRFs) \cite{lafferty2001conditional, mccallumli2003early}.  Such approaches were usually relying on fixed word embeddings and limited length observation windows over tokens (a single words or subword units in the input text) for feature engineering, as well as decision trees \cite{szarvas2006multilingual} or a set of binary Support Vector Machines (SVMs) \cite{mcnamee2002entity} on the part of the learning task. A typical example of semi-supervised sequence modeling approach for NER is \cite{chiu2016named}, where the K-Nearest Neighbors (KNN) classifier is used for pre-labeling of tweet data, after which the CRF model is applied in the sequential manner in order to produce the final predicted labels sequence.

In order to capture non-trivial long-distance dependencies in word or token sequences, neural networks capable of processing variable length inputs, like the recurrent neural network (RNN) and the long-short term memory (LSTM) units with the forget gate, were applied to NER classification tasks, which brought a significant performance improvement over the previous approaches \cite{chiu2016named}. Most recently, the concepts of bi-directional LSTMs and CNNs that learn both character- and word-level features were further improved with the introduction of pre-trained transformer based bi-directional representations provided by BERT type \cite{kenton2019bert} language models. Such contextualized language-model embeddings, comprising of token position, segment and token embedding are usually characterized as hybrid representations.

The key for development of cost effective solutions for different NLP tasks is ability to exploit learned representations of input data (inherently learned by LLM pre-training) and perform low-resource model adaptation in domain specific downstream tasks. It was made possible by recent advancements in self-supervised training of LLM architectures that are designed in the style of encoder, decoder or encoder-decoder deep neural networks (DNNs). BERT \cite{kenton2019bert} or bidirectional encoder representations from transformers are particularly well suited for NER task due to self-attention mechanism, which means that the encoder considers the entire context (e.g. in total up to 512 tokens for sentence, or multiple sentences in the paragraph) when predicting the category for a specific token, including observations from the past and future (i.e. both preceding and following tokens), due to its bidirectional training and structure. On the other hand, decoder type PTMs, like GPT \cite{brown2020language}, are generally considered as less suitable for NER and similar NLP tasks like sentiment analysis and masked word prediction,  due to unidirectional structure of decoder type PTMs that is well suited for word prediction and NLP tasks involving text generation, like text summarization, text completition or machine type translation. This was made possible by proposal of various learning strategies that have significantly improved representation learning by exploiting vast corpora of unannotated data. These include word-level objectives like causal language modeling (CLM) in \cite{brown2020language}, masked language modeling (MLM) in \cite{kenton2019bert} and  its span-level generalization in \cite{joshi2020spanbert}, replaced (token detection) language modeling (RLM) in \cite{Clark2020ELECTRA}, or denoising language modeling (DLM) in \cite{lewis2020bart}. Similarly, their sentence-level counter parts like next sentence prediction (NSP) in \cite{kenton2019bert}, sentence order prediction (SOP) in \cite{Lan2020ALBERT} or sentence contrastive learning (SCL) in \cite{kim2021self}. However, it should be noted that PTM performance varies depending on the type of downstream task, as well as the implementation, as suggested by \cite{ryu2021re}, where it was shown that BERT type \cite{kenton2019bert} baseline performs better in comparison to ALBERT model \cite{Lan2020ALBERT} on NER tasks, despite improvements that were brought by \cite{Lan2020ALBERT} over \cite{kenton2019bert} (e.g. lower memory consumption and increased training speed, without the NSP strategy \cite{liu2019roberta}).

When it comes to PTM based solutions for Serbian NLP, besides BERTić \cite{ljubevsic2021bertic} and its derivatives for QA \cite{cvetanovic2023} and NER \cite{ljubesic2023}, notable works relying on other PTMs also include learned embedding models proposed in \cite{milutin2024} and \cite{zivanic2024}. Significant efforts were also put into Serbian specific NER solutions proposed in \cite{perisic2023sr ,todorovic2021serbian}, while \cite{ikonic2024bert} considers the problem of using Serbian specific BERT based PTMs instead of multilingual BERTs \cite{conneau2020unsupervised, kenton2019bert} or south Slavic BERT models \cite{ljubevsic2021bertic}. As pointed out by \cite{skoric24modeli}, in the recent period there have been several attempts of developing Serbian specific PTMs like the \cite{Jerteh355} PTM based on RoBERTa architecture\cite{liu2019roberta}. However, when it comes to downstream tasks, according to \cite{skoric24modeli} and \cite{ikonic2024bert}, NER solutions based on Serbian specific PTMs achieve similar performance to NER models fine-tuned on BERTić \cite{ljubevsic2021bertic}, as measured by NER experiments involving seven entities: demonyms (DEMO), professions and titles (ROLE), works of art (WORK), person names (PERS), places (LOC), events (EVENT) and organizations (ORG); defined in SrpELTeC dataset proposed in \cite{todorovic2021serbian}.

\section{Methods and data}
\label{Methods and dataset}

Specific challenges for NER in legal documents usually stem from regulatory and compliance complexities, which come from the unique set of laws and regulations in each country, besides the language barrier that can be regarded as significant problem in the case of PTMs adaptation for languages with low NLP resources. From the literature it is known that the pre-training of BERT type models on the domain specific corpora (by adaptation or from scratch) can bring certain performance gains, like in the case of Legal-BERT \cite{chalkidis2020legal}. However, in practice such approach can be unfeasible for resource constrained solution designs. Similarly, our approach is more baseline in comparison to \cite{jin2023teamshakespeare}, where the Legal-LUKE model was trained from the beginning to solve MLM and NER at the same time (predict both words and entities, i.e. achieve legal-contextualized and entity-aware representations)

In this work we have decided to rely on the existing PTM proposed in \cite{ljubevsic2021bertic} and focus on the development of small scale dataset of legal texts that would allow us to demonstrate feasibility of fine-tuning BERTić \cite{ljubevsic2021bertic} model for the specific downstream task of recognizing 8 named entities (NEs) in Serbian legal documents. Defined NEs include: names of the courts (COURT), calendar dates (DATE), final rulings on the matter at hand (DECISION), names and abbreviations of written laws issued by the government (LAW), money values (MONEY), names of the announcements regarding new laws, amendments, and regulations exclusively from the official gazette of the Republic of Serbia (OFFICIAL GAZETTE), full names of the persons and annonimyzed abbreviations (NAME), alphanumeric designations of the court rulings (REFERENCE). The number and type of defined NEs is similar to legal NER solution described in \cite{ccetindaug2023named}.

Adopted approach can be regarded as similar to methods proposed in \cite{scherbakov2022finetuning, kovsprdic2024zero}, in the sense that the goal was to design effective NER system with the small amount of training data for the fine-tuning of the existing PTM. However, in comparison to \cite{scherbakov2022finetuning} proposed dataset is fully annotated, and we avoid training on partially annotated legal documents.

\subsection{Dataset development}
\label{Dataset development}

Legal texts, and especially court rulings, are known to contain long, nested and syntactically complex sentences, which are made of formal and domain specific language, with presence of complex abbreviations and cross-referencing on other documents. Although the process of digitization is ubiquitous, including optical character recognition (OCR) of the old archives, there is still a lack of annotated legal datasets in Serbian language, and a challenge of constantly evolving legal frameworks. Additional challenge is also that NEs in legal documents can contain synonyms, abbreviations, and misspellings.

\begin{figure}[!htb]
\centering
  \begin{subfigure}{\columnwidth}
    \centering
    \includegraphics[width=\columnwidth]{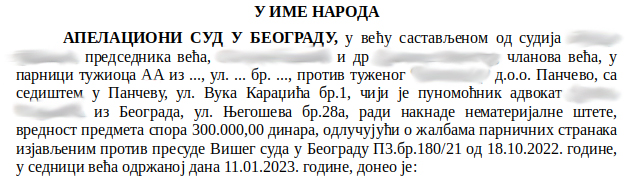}
    \caption{}
    \label{fig:figure_court_ruling}
  \end{subfigure}\vspace{1em}
   \begin{subfigure}{\columnwidth}
    \centering
    \includegraphics[width=\columnwidth]{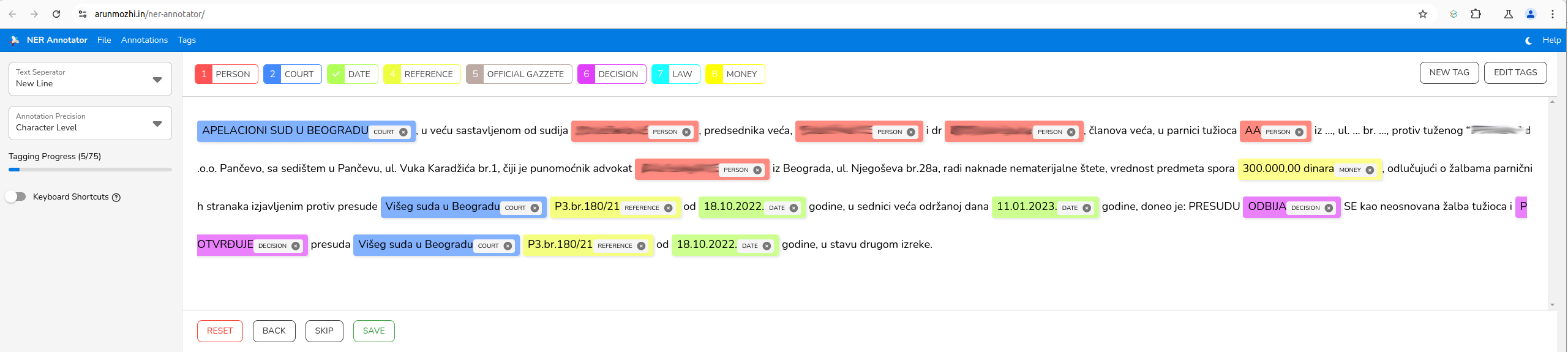}
    \caption{}
    \label{fig:figure_annotator}
  \end{subfigure}
   \begin{subfigure}{\columnwidth}
    \centering
    \includegraphics[width=\columnwidth]{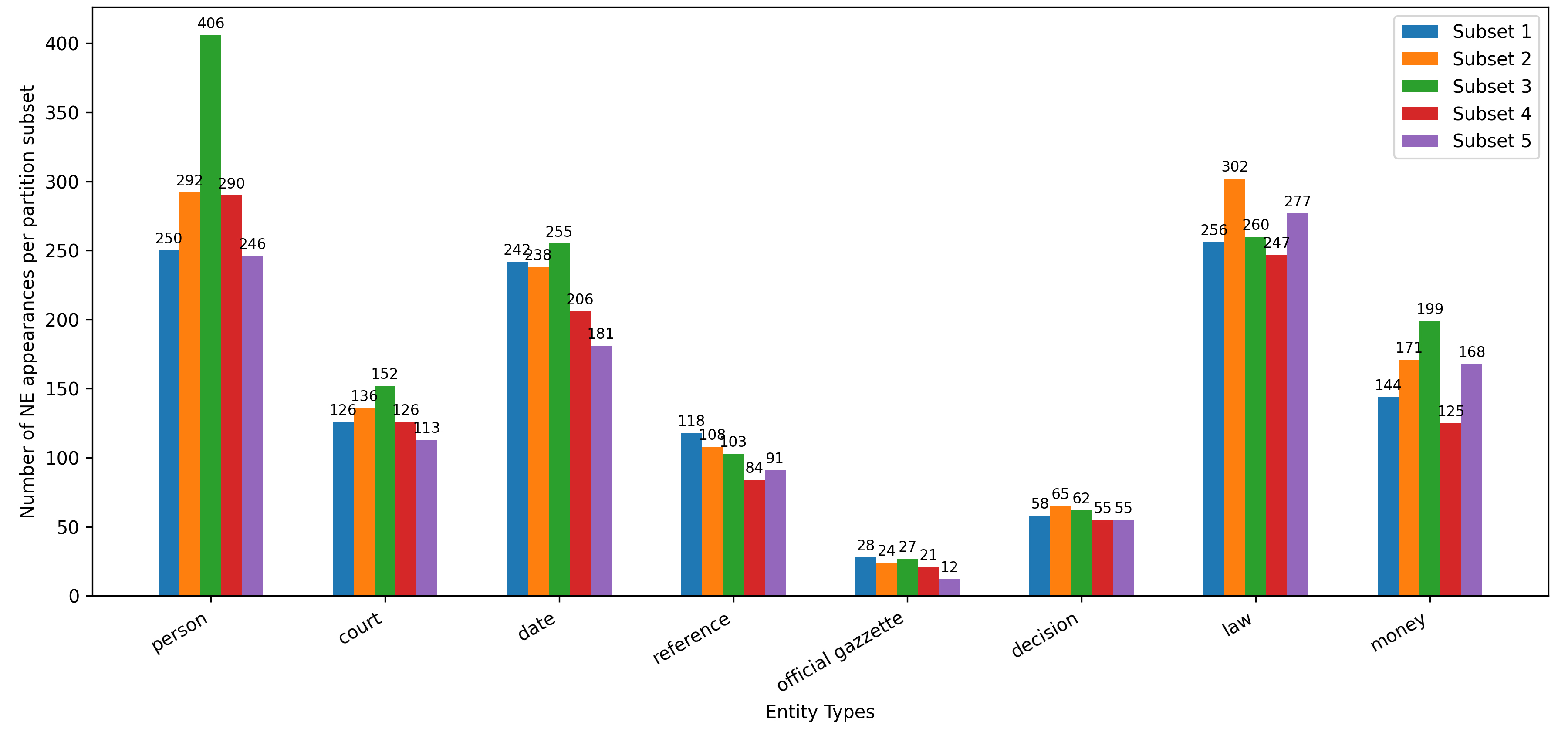}
    \caption{}
    \label{fig:figure_entity_count_distribution}
  \end{subfigure}
 %\captionsetup{width=0.98\linewidth}
    \captionsetup{justification=justified}
\caption{(a)~An illustration of original court ruling in Cyrillic script;  (b)~annotation process with BIO scheme; (c)~number of NE types appearances per each cross-validation subset (random sampling procedure is described in Algorithm~\ref{alg:expsetup}).}
\label{fig:NER dataset illustration}
\end{figure}

In order to most efficiently cover all types of legal documents in which previously described 8 NEs appear, we have decided to exclusively focus on currently available examples of Serbian judicial practice, which are available from the official website \cite{sudskaPraksa2024} of the Ministry of Justice. Such decision was motivated by the fact that this repository contains official public documents (court rulings) that were carefully chosen by the repository editors in order to provide representative samples of Serbian judical practice in appellate courts. Since these rulings, Fig.~\ref{fig:figure_court_ruling}, are mostly uniform in structure and cover appellations from municipal and high courts in Serbia, present NEs are quite diverse in both their values and context appearances. In contrast to Legal-BERT \cite{chalkidis2020legal}, we have not considered official law texts, which e.g. do not contain person names and court ruling references, while other NEs usually do not appear in the similar context as in the court rulings.

In total 75 documents addressing non-economic damages were collected from \cite{sudskaPraksa2024}. Since Serbian language can be written in both Cyrillic and Latin alphabet, and while BERTić \cite{ljubevsic2021bertic} PTM is producing more tokens in tokenization process of Cyrillic texts, for the purpose of NER design demonstration we have decided to perform transliteration of original documents from Cyrillic to Latin script. Since BERT type models are usually accepting up to 512 token length for encoder input, original documents were processed into one sentence per line textual files. Such data were manually annotated by using the tool available in \cite{arunmozhi2024}, Fig.~\ref{fig:figure_annotator}, with character level annotation precision (entity selection without white-spaces at the ends and punctuation marks) and outputs saved in JSON file format.

In the case of entities that consist of several words, there are different strategies for NEs annotation \cite{keraghel2024recentadvancesnamedentity}, e.g.  IO, BIO (or IOB), IOE, IOBES, IE, BIES (B(begin)- and  E(end)-, mark the first and the last token of an entity, S(single)- single token entity, while I(inside)- and  O(outside)- mark the tokens within and outside of the entities, respectively). it means that the number of instances of O category  (non-entity words) dominates the dataset. Although IO scheme was found to outperform others \cite{alshammari2021impact}, it comes with disadvantage of not being able to distinguish between consecutive NEs, while the most common BIO performs similarly to others and was therefore applied,  Fig.~\ref{fig:figure_annotator}. Note that sentences without any NE were not taken into final annotated dataset. Thus, annotation of 75 legal documents resulted in total 2172 sentences containing  NEs, i.e. 758012 characters which have produced 183543 tokens after applied WordPiece tokenization \cite{wordPieceTokenizer2016}, for which the same \cite{HuggingFaceTokenizer2024} tokenizer implementation as in selected PTM \cite{ljubevsic2021bertic} was utilized. Regarding the NE distribution, there were: 1484 "person", 653 "court", 1122 "date", 504 "reference", 112 "official gazette", 295 "decision", 1342 "law", and 807 "money" appearances, in total 6319. After application of BIO annotation scheme, instead of initial 8+1 entity types (8 described NEs and additional O-type) multi word NEs in the dataset produced in total 14+1 classification categories ("date" and "decision" NE do not produce an I-type token label).

\subsection{Model development}
\label{Model development}

Adopted PTM was pre-trained using ELECTRA framework \cite{Clark2020ELECTRA}, i.e.  replaced token detection language modeling (RLM), which is characterized by the following optimization objective:

\begin{equation}
{\cal{L}}_{\text{PTM}} = {\cal{L}}_{G} + \lambda {\cal{L}}_{D} ,
\label{eq:PTM_objective}
\end{equation}
where ${\cal{L}}_{G}$ denotes the standard MLM pre-training of generative BERT encoder $G$, while ${\cal{L}}_{D}$ is the objective of the  RLM specific discriminator $D$. If we denote with $\hat{x}_{t}$  masking of token $t$ in MLM, and with $\tilde{x}_t$ token replacement with the word that was generated by $G$ in RLM, then the corresponding objectives can be defined by probabilities $P_G$ and $P_D$,  which correspond to probability of generator $G$ correctly predicting the masked word ${x}_{t}$ in sequence $\textbf{x}$, or probability of discriminator $D$ correctly detecting that the original word ${x}_{t}$ was replaced by the generated $\tilde{x}_t$:

\begin{equation}
{\cal{L}}_{G} = -\sum_{t \in M} y_t \log P_G(x_t | \hat{x}_{t}) ,
\label{eq:PTM_objective_G}
\end{equation}

\begin{equation}
\begin{aligned}
{\cal{L}}_{D}  = -\sum_{t=1}^{T} \bigl[ & y_t \log P_D(y_t = 1 | \tilde{x}_t) \\&+ (1 - y_t) \log P_D(y_t = 0 | \tilde{x}_t) \bigr],
\end{aligned}
\label{eq:PTM_objective_D}
\end{equation}

\noindent where $y_t=1$ indicates the word replacement or masking operation.

We note that $G$ is not trained in an adversarial fashion, characteristic for generative adversarial networks (GANs), but separately from $D$, while ${\cal{L}}_{D}$ contains two terms encompassing discriminator decisions in both cases (when the original word $x_t$ in the sequence is present at the input of $D$, as well as when it is replaced by the MLM generator $G$), which increases the number of tokens $T$ contributing to model update. This approach makes training more efficient, as the PTM learns from every token rather than just masked ones. Pre-trained discriminator $D$ is then fine-tuned on sentences from legal documents for token classification task with 15 unique labels (categories) corresponding to described NER in Section~\ref{Dataset development}.

\subsection{Experimental setup}
\label{Experimental setup}

In order to effectively perform and assess proposed downstream adaptation of PTM, the following statistical cross-validation procedure was used. Initially created dataset with imbalanced number of tokens per NE types was first randomly shuffled at document level into five disjunct subsets, i.e. random 5-fold cross-validation partition, as shown in Fig.~\ref{fig:figure_entity_count_distribution}. It is achieved by iterative clustering procedure described in Algorithm~\ref{alg:expsetup}, consisting of grouping of documents with similar NE distributions:

\begin{algorithm}
\caption{}
\begin{algorithmic}[1]
    \Procedure{\textsc{Training/test set random partition}}{}
        \State \textbf{\textit{Initialization:}} \textit{${\cal{K}}$ annotated documents, with ${\cal{C}}$ NE types (categories); set number of partition ${\cal{P}}$ subsets $K$, ${\cal{P}}=\{S_1, ..., S_K\}$}
         \State \textbf{\textit{Iterative procedure:}}
        \State \textbf{\textit{\# 1:}} \textit{Assign feature vector $f_i$ to each document $D_i$, reflecting the NE content in labeled sentences, e.g. normalized histogram of NE appearances in $D_i$}
        \State \textbf{\textit{\# 2:}}~\textit{Group $D_i$, $ i=1..{\cal{K}}$, into $K$ subsets $\tilde{S}_k$ based on similarity in feature space $f_i\in\mathbb{R}^{\cal{C}}$, e.g. by $K$-means clustering with $L_p$ distance}:\\
        \hspace{4em}$\min_{\{\tilde{S}_1, \tilde{S}_2, \dots, \tilde{S}_K\}} \sum_{k=1}^{K} \sum_{f_i \in \tilde{S}_k} \|x_i - \mu_k\|_p^p$
        \State \textbf{\textit{\# 3:}} \textit{By sampling without replacement, randomly select one document per each $\tilde{S}_k$ and assign that document to final partition subset $S_k$. If some of $\tilde{S}_k$ become empty, continue the iterative procedure over remaining ones until all $D_i$, $ i=1..{\cal{K}}$ are assigned  to some ${S}_k\in{\cal{P}}$.}
        \State \textbf{\textit{Final Deduplication:}} \textit{Remove duplicates, i.e. identical sentences that appear in all documents in the same partition subset $S_k$ (e.g. name of the court in the heading)}
    \EndProcedure
\end{algorithmic}
    \label{alg:expsetup}
\end{algorithm}

Although random partition of original dataset is performed on document level, final training/test subsets are made only of one sentence per line entries from corresponding documents. Applied procedure with parameters ${\cal{K}}=75$, ${\cal{C}}=9$, $K=5$, $L_1$ distance and NE type histograms as feature vecors $f_i$  resulted in partition shown in Fig.~\ref{fig:figure_entity_count_distribution}. Class "O" of tokens denoting words outside 8 predefined NEs expectedly had several orders of magnitude higher number of instances and therefore was not shown on the same diagram in Fig.~\ref{fig:figure_entity_count_distribution}, although it was taken into account for $f_i$ computation. Note that proposed randomization procedure did not consider final 14+1 categories of NEs that are obtained after text tokenization, as described in Section~\ref{Dataset development}.

\subsection{Model training process}
\label{Model training}

\begin{figure}[!b]
\centering
  \begin{subfigure}{0.94\columnwidth}
    \centering
    \includegraphics[width=0.8\columnwidth]{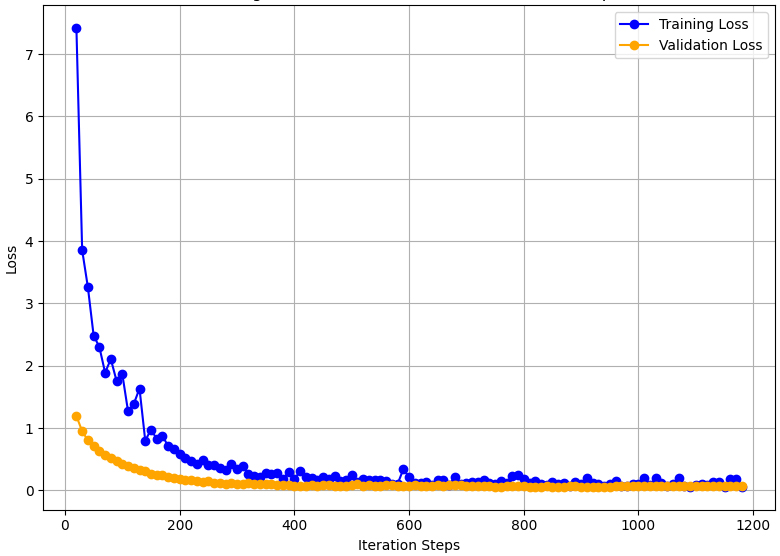}
    \caption{}
    \label{fig:figure_loss_vs_iteration_steps}
  \end{subfigure}\vspace{0.6em}
    \begin{subfigure}{0.94\columnwidth}
    \centering
    \includegraphics[width=0.8\columnwidth]{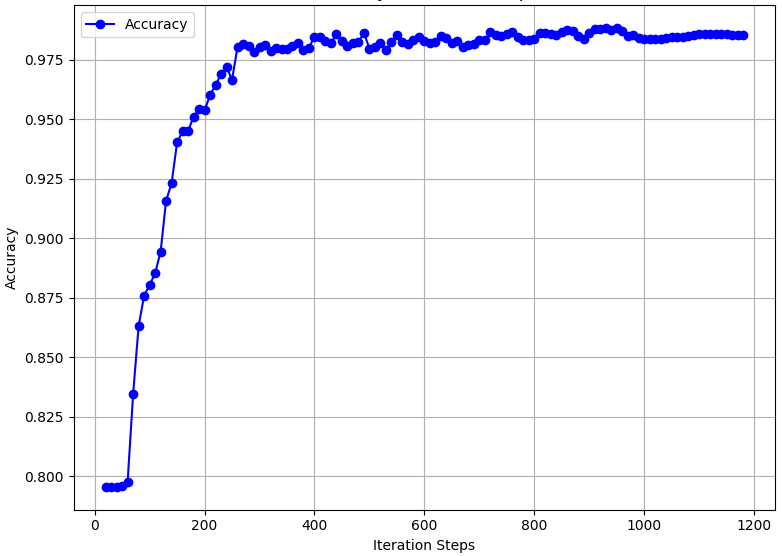}
    \caption{}
    \label{fig:figure_accuracy_vs_iteration_steps}
  \end{subfigure}
   \captionsetup{width=0.98\linewidth}
\caption{(a)~Model optimization loss, and (b)~mean accuracy over training iteration steps.}
\label{fig:Model training convergence}
\end{figure}

Model performance was measured by 5 independent experiments involving selection of 4 partition subsets for training and 1 for test. Production model parameters were obtained by training over all available data. Model training convergence over one of the cross-validation folds is illustrated in Fig.~\ref{fig:Model training convergence}.

\begin{figure}[!b]
\centering
\begin{subfigure}{0.98\columnwidth}
    \centering
    \includegraphics[width=0.98\columnwidth]{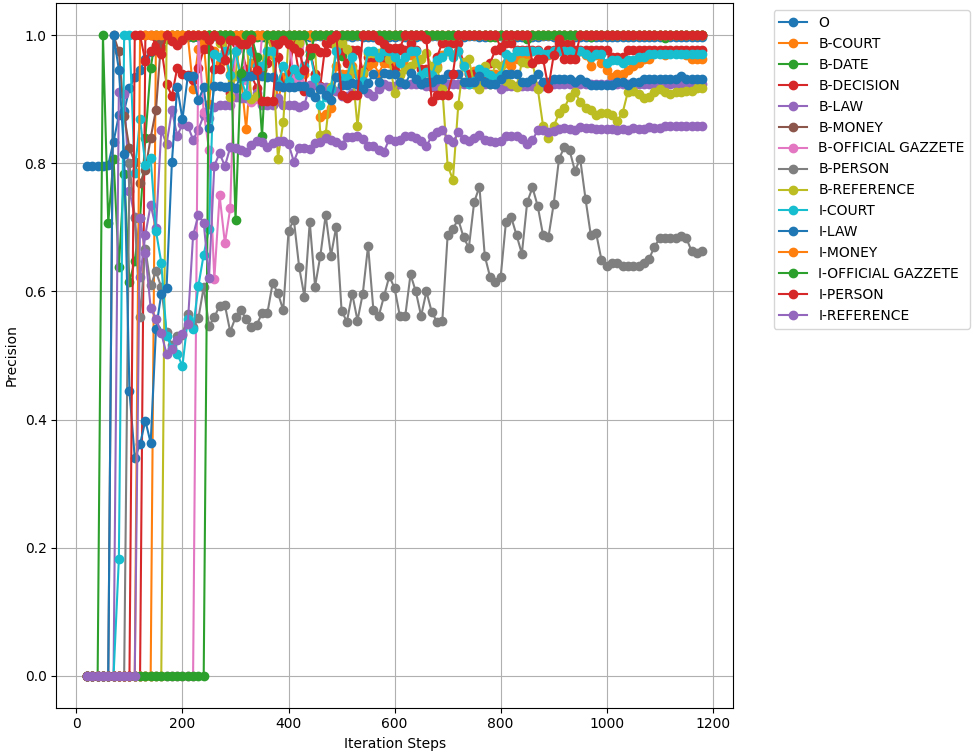}
    \caption{}
    \label{fig:figure_precision_entities_vs_iteration_steps}
  \end{subfigure}\vspace{0.6em}
 \begin{subfigure}{0.98\columnwidth}
     \centering
    \includegraphics[width=0.98\columnwidth]{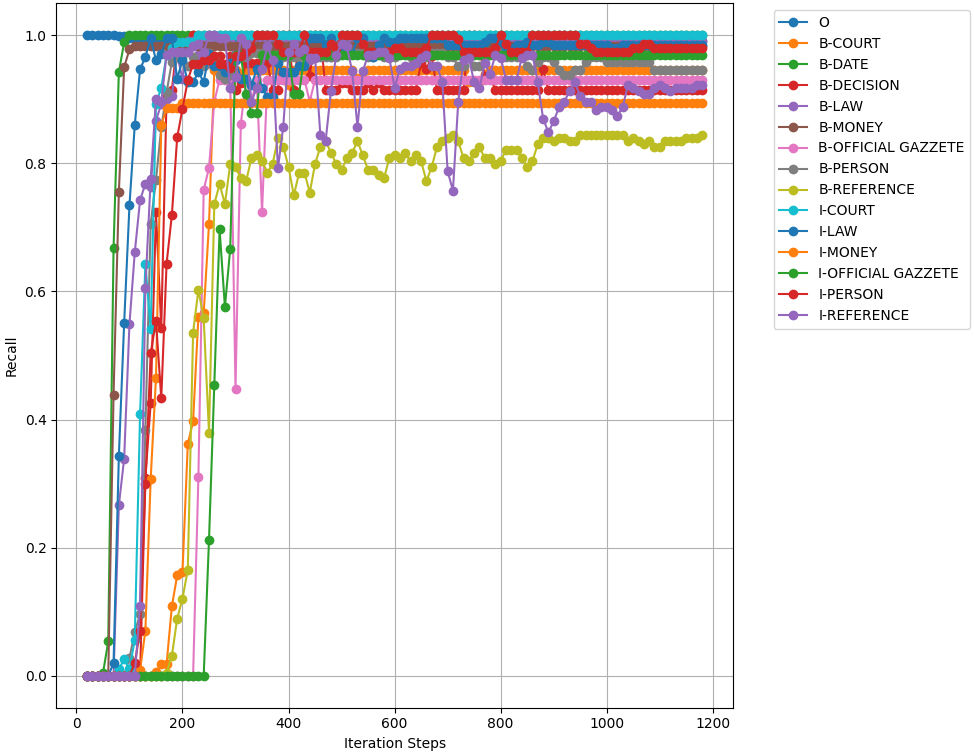}
    \caption{}
    \label{fig:figure_recall_entities_vs_iteration_steps}
  \end{subfigure}
 \captionsetup{width=0.98\linewidth}
\caption{(a)~Precision and  (b)~recall curves for each of 15 NER output classes (categories) over training iteration steps.}
\label{fig:Training precision and recall}
\end{figure}

For model implementation was used Python programming language and cloud computing platform with one A100-SXM4 GPU device and 40GB of memory. Each of the 5 training sessions consisted of selecting 4 out of 5 partition subsets and random splitting of their labeled sentences into temporary training and validation sets in the ratio 9:1, repsectively. After initial pre-study involving AdamW optimizer, transformer DNN model parameters were optimized using learning rate of 2e-5, weight decay 0.01, fp16 precision, 4 gradient accumulation steps, number of training epochs 6, and the best model selection based on the mean validation set $F_1$ score. Since the input sequence length was limited to 512 tokens, training batch size was deliberately set to 2 in order to avoid any possibility of exceeding the maximum input length in case of long and nested sentences in legal documents, which can be regarded as suboptimal solution. Number of labeled NE training samples varied between 1143 and 1464 per training session. Since the total number of the labeled sentences was 2172, approximately 90\% of 1738 was used in each training session. This corresponds to $\approx1560$ sentences for each training epoch, with the effective batch size of 8 (due to 4 gradient accumulation steps), which leads to around $\approx195$ training steps per epoch, i.e. $\approx1172$ iterations. A more detailed illustration of the training process over each of the 15 NER model output categories (BIO labels) is shown in Fig.~\ref{fig:Training precision and recall}. The number of steps in Fig.~\ref{fig:Model training convergence} and Fig.~\ref{fig:Training precision and recall} is the same, but in general varies between the experiments.

\begin{figure}[!htb]
\centering
    \includegraphics[width=0.98\columnwidth]{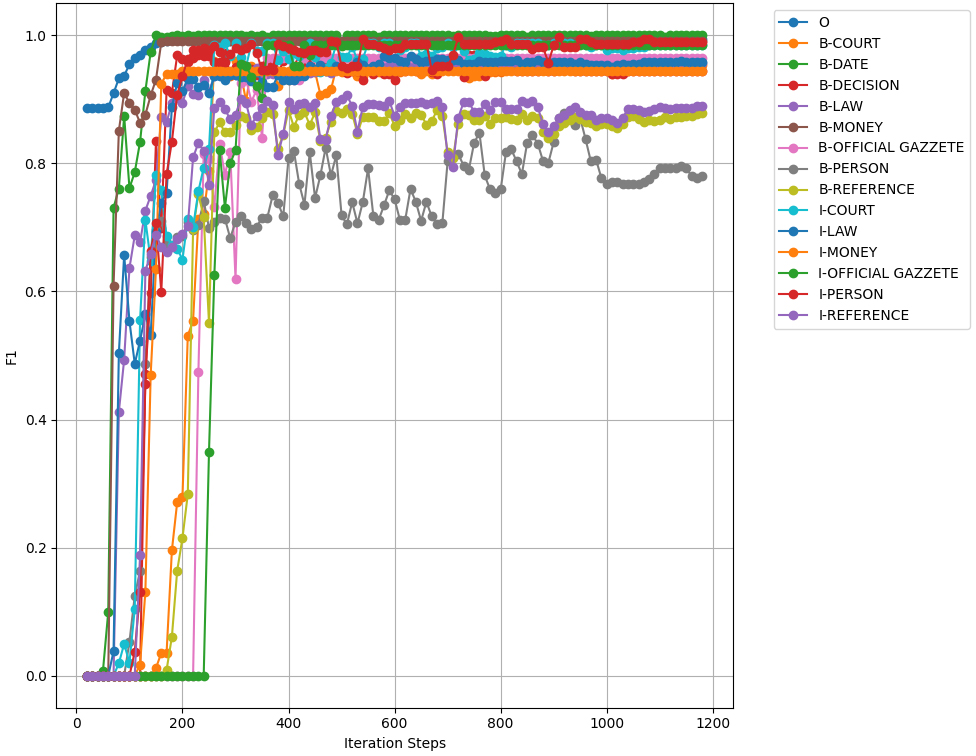}
   \captionsetup{width=0.98\linewidth}
\caption{$F_1$ measure per each output class vs  training iterations.}
\label{fig:figure_f1_entities_vs_iteration_steps}
\end{figure}

We note that precision is measured as the percentage of labeled NEs found by the NER system that are correct, while the recall is the percentage of labeled NEs present in the corpus that are found by the system. A NE was considered as the correct only if it was an exact match of the corresponding entity in the data file. Individual $F_1$ measures over each of the classes during training are shown in Fig.~\ref{fig:figure_f1_entities_vs_iteration_steps}

Final production model had 450 MB in size, 110 M trainable parameters and its training using all available data lasted under ten minutes on the given hardver. During training total memory reserved on the GPU did not exceed 4 GB, while the initially allocated memory was around 1.4 GB in size.

\section{Results and discussion}
\label{Results and discussion}

Proposed NER solution was extensively tested using objective performance measures, as well as by additional model analysis involving human judgment of NER quality and effectiveness. Model efficiency was not specifically analyzed, but the model size was such that it was easy to run the production model on local notebook machine. Subjective evaluation was done in order to test developed solution on out of sample sentences and investigate model robustness against different kinds of noisy inputs. More details regarding obtained results and their analysis is provided in the following.

\subsection{Cross-validation performance}
\label{Cross-validation performance}

\begin{figure}[tb]
  \centering
  \includegraphics[width=\columnwidth]{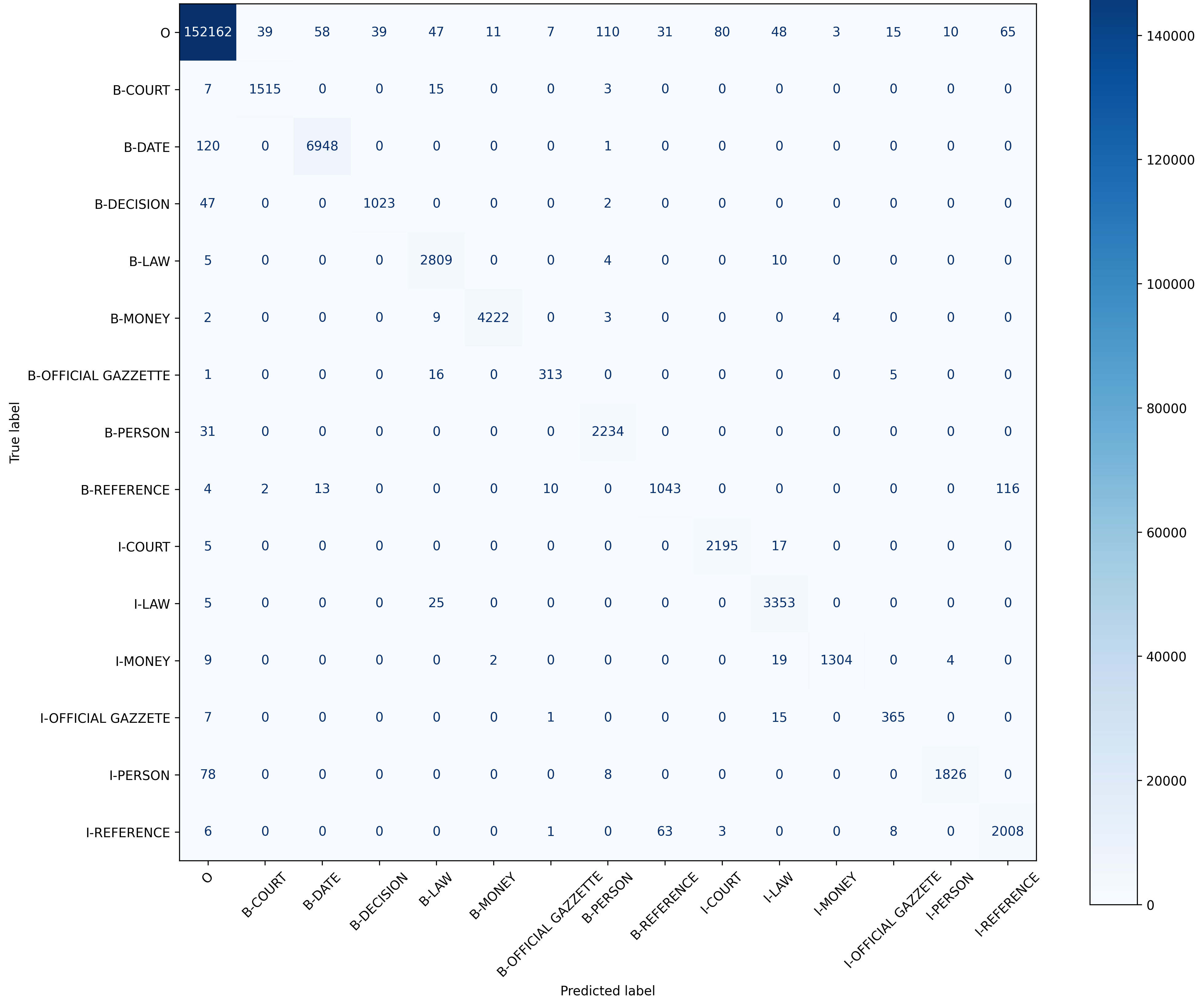}
  \captionsetup{justification=justified}
\caption{Accuracy assessment matrix}
\label{fig:Acuracy assesment matrix}
\end{figure}

Detailed statistical cross-validation of model performance was performed according to procedure outlined in Section~\ref{Model training}, which consisted of estimation of accuracy ($Acc$), precision ($Prec$), recall ($Rec$) and $F_1$ measure. Besides individual metrics computed for each NER category (${\cal{C}}=15$ output classes), their statistical averages were also reported, the last row in Table~\ref{tab:performance_metrics}. All metrics were computed based on results of 5 independent test experiments, in which there was no overlap between the legal documents in the training and test sets. Individual NER results are aggregated and shown in accuracy assessment matrix in Fig.~\ref{fig:Acuracy assesment matrix}.

The average values in Table~\ref{tab:performance_metrics} were computed by macro averaging, in which the metrics were first calculated individually for each class based on values in Fig.~\ref{fig:Acuracy assesment matrix} and then averaged across all classes. Alternative aggregate metrics would be the ones obtained by micro averaging that would compute number of true positives ($TP$), true negatives ($TN$) and false negatives ($FN$) across all classes and then compute aggregate metrics by their averaging. However, this approach was not pursued since it would lead to more optimistic aggregate metrics due to bias that would be introduced by class "O" with several orders of magnitude more samples. Similarly also holds for $Acc$, which is also known to have overly optimistic value in case of large number of negative samples that are correctly classified. Thus, based on results in Fig.~\ref{fig:Acuracy assesment matrix} the reported metrics were computed by following expressions:

\begin{equation}
\overline{Prec}= \frac{1}{{\cal{C}}} \sum_{i=1}^{{\cal{C}}} \frac{{TP}_i}{{TP}_i + {FP}_i},\\
\label{eq:precision}
\end{equation}

\begin{equation}
\overline{Rec} = \frac{1}{{\cal{C}}} \sum_{i=1}^{{\cal{C}}} \frac{{TP}_i}{{TP}_i + {FN}_i},\\
\label{eq:recall}
\end{equation}

\begin{equation}
\overline{F1} = \frac{1}{{\cal{C}}} \sum_{i=1}^{{\cal{C}}} \frac{2 \cdot {Precision}_i \cdot {Recall}_i}{{Precision}_i + {Recall}_i},
\label{eq:f1_measure}
\end{equation}

\begin{equation}
{Acc} = \frac{1}{{\cal{C}}} \sum_{i=1}^{{\cal{C}}} \frac{TP_i + TN_i}{TP_i + TN_i + FP_i + FN_i}.
\label{eq:accuracy}
\end{equation}

\subsection{Model robustness}
\label{Model robustness}

\begin{figure*}[!t]
    \centering
    \begin{subfigure}[b]{0.3\textwidth}
        \includegraphics[width=\textwidth]{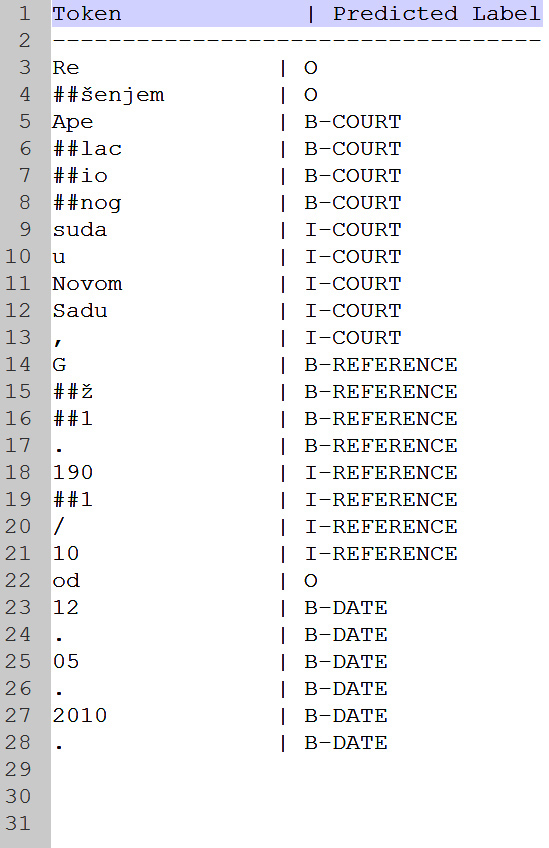}
        \caption{}
        \label{fig:NER_examples_noisy_comparison_A}
    \end{subfigure}
    \hfill
    \begin{subfigure}[b]{0.3\textwidth}
        \includegraphics[width=\textwidth]{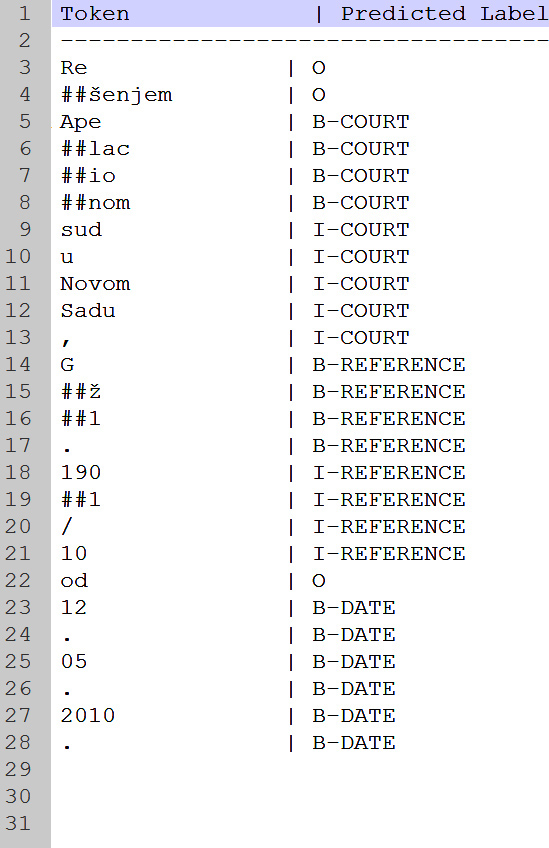}
        \caption{}
        \label{fig:NER_examples_noisy_comparison_B}
    \end{subfigure}
    \hfill
    \begin{subfigure}[b]{0.3\textwidth}
        \includegraphics[width=\textwidth]{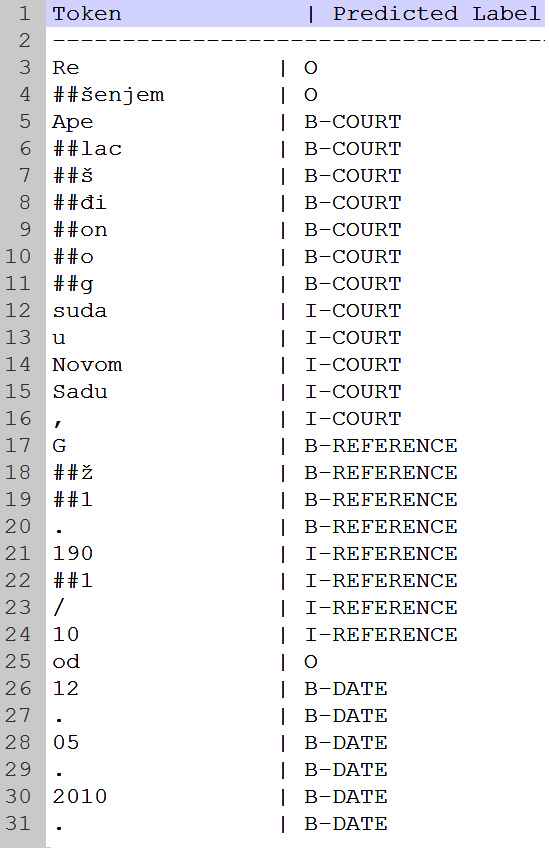}
        \caption{}
        \label{fig:NER_examples_noisy_comparison_C}
    \end{subfigure}
    \caption{Proposed NER in the case of noisy inputs: (a)~regular text, (b) and (c)~text with the presence of errors and misspellings; in all cases the outputs are correctly classified.}
    \label{fig:figure_NER_examples_noisy_comparison}
\end{figure*}

After successful cross validation experiments which have confirmed initial research hypothesis that it is possible to create effective domain specific NER solution for Serbian language with small amount of resources, we have also investigated solutions's robustness. E.g. in Fig.~\ref{fig:figure_NER_examples_noisy_comparison} are shown some illustrations of NER outputs (token level classification decisions), which are correct despite the presence of text errors and misspellings.

\subsection{Results analysis}
\label{Results analysis}
Values in Table~\ref{tab:performance_metrics} suggest that model in general performs well, although slightly weaker on "B-Reference" and "I-Reference" classes. It could be expected due to diverse nature of "reference" NE values, which can contain various alphanumeric characters, spacings, syntax that varies among different courts and usually consist of character sequences that do not generally correspond to any word from the language. On the other hand, the proposed dataset had relatively limited corpus of annotated examples (e.g. 504 "reference" NE), implying that the more complex NEs would be harder to learn by limited amount of training data. Slightly better are the results for "official gazette" NE, which according to Fig.~\ref{fig:figure_entity_count_distribution} had even smaller corpus of 112 annotated samples over whole training/test set partition. However, NER model adaptation based on PTM was still successful, since these NEs had smaller variability in comparison to previously discussed "reference" type.
Results of model training convergence in Fig.~\ref{fig:figure_precision_entities_vs_iteration_steps} reveal that improvements of model precision on "person" NER are relatively slow and underperforming in comparison to other NEs. On the other hand, recall values change in similar fashion to other NEs, Fig.~\ref{fig:figure_recall_entities_vs_iteration_steps}. This indicates that recognition of "person" NE is almost always successful in case of real or correct words corresponding to person names, but there is also a large number of misclassification or false positives (see. e.g. $F_1$ measure in able~\ref{tab:performance_metrics}). Although this NE is the most frequent in the created dataset, Fig.~\ref{fig:figure_entity_count_distribution}, observed model behaviour could be due ti the fact that "person" NE also encompass name abbreviations in the form of initials (e.g. "AA", "CC", ...), which correspond to anonymized personal data in public court rulings. Thus, possible improvements could be directed towards these kind of challenges.

\begin{table}[tb]
    \centering
    \caption{Performance metrics: per class and average values.}
    \begin{tabular}{p{1.6cm}|p{1.2cm}|p{1.2cm}|p{1.2cm}|p{1cm}}
        \toprule
        Class & $Recall$ & $Precision$ & $Accuracy$ & $F_1$ \\
        \midrule
        O & 0.99 & 0.99 & 0.99 & 0.99 \\
        B-Court & 0.98 & 0.97 & 0.99 & 0.97 \\
        B-Date & 0.98 & 0.98& 0.99 & 0.96 \\
        B-Decision & 0.95 & 0.96& 0.99 & 0.95 \\
        B-Law & 0.99 & 0.96 & 0.99 & 0.97 \\
        B-Money & 0.99 &0.99  & 0.99 & 0.99 \\
        B-O.Gazette & 0.93 & 0.94 &0.99 & 0.93 \\
        B-Person& 0.98 & 0.94 & 0.99 & 0.96 \\
        B-Reference & 0.87& 0.91& 0.99 & 0.89 \\
        I-Court & 0.99 & 0.96  & 0.99 & 0.97 \\
        I-Law & 0.99 &0.96 & 0.99 &0.97 \\
        I-Money &0.97 & 0.99  & 0.99 & 0.98 \\
        I-O.Gazette &0.94 &  0.92 & 0.99 & 0.93 \\
        I-Person & 0.95 & 0.99 & 0.99 & 0.97 \\
        I-Reference & 0.96 & 0.91 & 0.99 & 0.93 \\
        \hline
        Average & \textbf{0.97} &\textbf{0.96} &\textbf{0.99} & \textbf{0.96}\\
        \bottomrule
    \end{tabular}
    \label{tab:performance_metrics}
\end{table}

\section{Conclusions}
\label{Conclusion}
In the presented work we have demonstrated design of novel NER system for Serbian legal documents and proposed novel domain specific dataset based on publicly available court rulings. Besides model development and conducted experiments, the paper also addresses the methodology of successful adaptation of pre-trained language models for specific downstream NLP tasks. This is particularly important in case of languages and applications with low resources, in terms of NLP tools and specific training corpora. We hope that this work will stimulate further interest into this emerging topic, especially in the case of Serbian language.

\section*{Acknowledgment}
The authors would especially like to acknowledge previous collaboration with other AI4Legal team members: attorney at law Sanja Kalušev and M.Sc. Milica Brković, with whom AI~ACTA project proposal was made in 2023.

The second author would also like to acknowledge support by the Science Fund of the Republic of Serbia through the grant agreement no. 7449, project AI-SPEAK - "Multimodal multilingual human-machine speech communication", and the Ministry of Science, Technological Development and Innovation of the Republic of Serbia (Contract No. 451-03-137/2025-03/200156) for the support by the project “Scientific and artistic research work of researchers in teaching and associate positions at the Faculty of Technical Sciences, University of Novi Sad 2025” (No. 01-50/295).

%\section*{References}

%
%Please number citations consecutively within brackets \cite{b1}. The
%sentence punctuation follows the bracket \cite{b2}. Refer simply to the reference
%number, as in \cite{b3}---do not use ``Ref. \cite{b3}'' or ``reference \cite{b3}'' except at
%the beginning of a sentence: ``Reference \cite{b3} was the first $\ldots$''
%
%Number footnotes separately in superscripts. Place the actual footnote at
%the bottom of the column in which it was cited. Do not put footnotes in the
%abstract or reference list. Use letters for table footnotes.
%
%Unless there are six authors or more give all authors' names; do not use
%``et al.''. Papers that have not been published, even if they have been
%submitted for publication, should be cited as ``unpublished'' \cite{b4}. Papers
%that have been accepted for publication should be cited as ``in press'' \cite{b5}.
%Capitalize only the first word in a paper title, except for proper nouns and
%element symbols.
%
%For papers published in translation journals, please give the English
%citation first, followed by the original foreign-language citation \cite{b6}.

\bibliographystyle{IEEEtranDOI}
%\footnotesize
\bibliography{NER4Legal_SRB_bibliography}

\end{document}